\documentclass[]{spie}  %>>> use for US letter paper/
 \usepackage{threeparttable}
 \usepackage{subcaption}
\usepackage{multirow}
\usepackage{threeparttable}
\usepackage{booktabs}

\usepackage{amsmath,amsfonts,amssymb}
\usepackage{graphicx}
\usepackage[colorlinks=true, allcolors=blue]{hyperref}

\title{PHT-BOT:DEEP-LEARNING BASED SYSTEM FOR AUTOMATIC RISK STRATIFICATION OF COPD PATIENTS BASED UPON SIGNS OF PULMONARY HYPERTENSION}

\author[a]{David Chettrit}
\author[a]{Orna Bregman Amitai}
\author[b]{Itamar Tamir}
\author[a]{Amir Bar}
\author[a]{Eldad Elnekave}
\affil[a]{Zebra Medical Vision Ltd}
\affil[b]{Rabin Medical Center}

\authorinfo{Send correspondence to D.C.
\newline
E-mail: chettrit.david@zebra-med.com 
}

\begin{document} 
\maketitle

\begin{abstract}
Chronic Obstructive Pulmonary Disease (COPD) is a leading cause of morbidity and mortality worldwide. Identifying those at highest risk of deterioration would allow more effective distribution of preventative and surveillance resources. Secondary pulmonary hypertension is a manifestation of advanced COPD, which can be reliably diagnosed by the main Pulmonary Artery (PA) to Ascending Aorta (Ao) ratio. In effect, a PA diameter to Ao diameter ratio of greater than 1 has been demonstrated to be a reliable marker of increased pulmonary arterial pressure. Although clinically valuable and readily visualized, the manual assessment of the PA and the Ao diameters is time consuming and under-reported. 
\par The present study describes a non invasive method to measure the diameters of both the Ao and the PA from contrast-enhanced chest Computed Tomography (CT). The solution applies deep learning techniques in order to select the correct axial slice to measure, and to segment both arteries. The system achieves test Pearson correlation coefficient scores of 93\% for the Ao and 92\% for the PA. To the best of our knowledge, it is the first such fully automated solution.
\end{abstract}


\section{INTRODUCTION}
\label{sec:intro}
Chronic Obstructive Pulmonary Disease (COPD) affects between 4-8\% of the western population. The disease course is variable, with some individuals virtually asymptomatic and others prone to repeated exacerbations and functional decline. Pulmonary hypertension (PHT) defines increased blood pressure within the pulmonary arteries. The main PA is anatomically positioned adjacent to the Ao. A PA to Ao ratio of greater than 1 has been demonstrated to be a reliable marker of increased pulmonary arterial pressure \cite{Iyer14,Corson14}. Enlargement of the PA or a PA to Ao ratio of greater than 1, was also shown to be among the most robust predictors of COPD exacerbation \cite{Adir14,Karakus15,Wells12}. This is clinically intuitive, as the presence of secondary pulmonary hypertension is a sign of advanced COPD. 
\par Pulmonary hypertension can be diagnosed via direct intra-vascular pressure measurements. It can also be inferred from radiographic findings, specifically enlargement of the pulmonary arteries, which can be rapidly assessed by comparing the diameters of the PA to Ao. This assessment of the Ao and PA diameter is time consuming and often overlooked by radiologists. Our system provides automatic measurements of the main PA and the Ao. In the COPD population, this metric may be utilized as input for population health risk-stratification. In general, these measurements may identify those who could benefit from further evaluation for pulmonary hypertension and, if the Ao is enlarged, screened for aortic aneurysms. The described solution has been developed for scans with intravenous contrast, where the blood vessels in the scan are enhanced, thus facilitating their measurements.

\begin{figure}[ht]
  \centering
   
  \includegraphics[width=.8\linewidth] {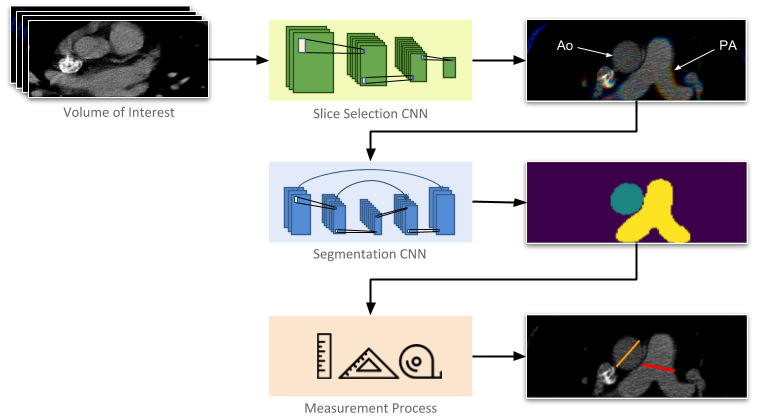}%system_flow_last %{figures/core_system_workflow.jpg}
  \captionsetup{margin=1cm}
  \vspace*{0.3cm}
    \caption{\label{fig:system_design_pht}The core system flow with its three consecutive steps of and their respective output. Given the volume of interest, the slice selection CNN is used to select the correct slice for the measurements. The two arteries are labelled in white in the chosen slice, for the reader's convenience. The chosen slice segmentation map is then generated using a second CNN. Finally, image processing techniques are used to measure the diameters, as shown in orange for the Ao and red for the PA.}
\end{figure}

\par The PA and Ao diameters are measured at the main PA bifurcation, on a unique axial slice. The PA is measured above its bifurcation at its maximum diameter, and the Ao is measured at the same level \cite{Ussavarungsi14,Lee15}. The system described here measures the arteries in this specific way in order to simulate as accurately as possible the radiologist's process.

\section{PREVIOUS WORK}
To the best of our knowledge, no deep learning based solution for fully automated measurement of the PA to Ao ratio on contrast chest CT scan has ever been presented. Previous work included semi-automated solutions based on traditional computer vision techniques to measure either both arteries, or just the PA diameter \cite{Linguraru10,Moses16}. Yiting et al. \cite{Yiting15} developed an automated solution on low-dose non contrast CT scan solution using an anatomy label map as a seed to segment the arteries of interests. The solution we propose was trained to precisely locate the correct axial slice for measurement, and to segment it automatically using a convolutional neural network (CNN). CNNs have become popular in natural image classification problems since Krizhevsky et al. \cite{Krizhevsky12} presented AlexNet, a convolutional neural network which outperformed previous traditional models on natural image classification problems. Since then, CNNs have been further developed and deeper architectures were introduced. Among others, UNet architectures for CNN introduced by Ronneberger et al. \cite{Ronneberger15} were designed for segmentation tasks, especially on bio-medical image segmentation tasks. These models were proven to be reliable, as well as efficient tools in medical imaging, as summarized by Shen and his team. \cite{Shen17} Wang et al. \cite{Wang17} demonstrated they could be used to automatically detect pathologies on medical X-ray, and Shin et al. \cite{Shin16} provided a similar proof on computed tomography scans. Laserson et al. \cite{Laserson18} also demonstrated the power of CNNs to detect 40 pathologies on chest X-rays. They also reported a higher rate of radiologist-algorithm agreement, than radiologist-radiologist agreement.

\section{Material}
\subsection{Data}
Altogether, 1,285 Chest CT studies were used for training and validation (81.7\% of our data set), the remaining 288 studies were used for testing (18.3\%). 600 Chest CT studies were manually annotated by three experienced radiologists. The radiologists measured both arteries diameters and also reported the measurement slice. The ground truth was set to the mean of the three radiologists measurements. Additionally, 685 studies with annotated masks per artery were used for the segmentation model training and validation.
For testing and reporting our results, we used an additional set unseen during training. In this set, the exact measurement locations were marked on the scans, in order to visualize the measurement process of the radiologist, and to compare it with ours.
Table \ref{table:data_distribution} presents the diverse properties of our data sets. In effect, this table includes statistics regarding the age, sex, size, as well as the ratio of cases with PA to Ao ratio greater than 1, within our training, validation and test sets. Our data set was labelled differently, since it is used for classification, segmentation and measurements tasks. The segmentation set, for example, does not contain ground truth measurements of both arteries, so no statistics are given regarding the distribution of cases with PA to Ao ratio $>$ 1 within this set.

\begin{table}
\setlength{\tabcolsep}{10pt}
\begin{center}
\begin{threeparttable}
\caption{Data properties distribution}\label{table:data_distribution}
\begin{tabular}{lccc}
\toprule
\toprule
& \multicolumn{2}{c}{ Train \& Validation}  & Test \\
% \rule{0pt}{3ex} 
\quad Data type & Measurement & Segmentation & Measurement \\
\midrule
\quad size (studies number) & 600 & 685 & 288  \\
\quad female ratio (\%) & 66 & 49 & 46  \\
\quad age mean (standard deviation) & 63.1 (19.7) & 66.0 (17.4) & 62.0 (16.8)  \\
\quad cases with PA to Ao $>$ 1 (\%) & 12.8 & -  & 9.0 \\
\bottomrule
\end{tabular}
\begin{tablenotes}[para,flushleft]
Distribution of various properties of interest inside the train, validation and test data sets: size, female ratio, age mean (with standard deviation) and ratio of cases with PA to Ao $>$ 1. The statistics on the train and validation sets are presented together in the first two columns denoting the type of data set (measurement or segmentation).
\end{tablenotes}
\end{threeparttable}
\end{center}
\end{table}

\subsection{Preprocessing}
Our system uses thresholding and connected components analysis, as well as morphological operations in order to detect the lungs, the trachea and the carina. Based on the location of the lungs and the carina, a volume of interest around the main PA bifurcation is defined.
This volume usually comprises between 20 to 50 slices, depending on the slice thickness and increment of the series. In order to focus on the two arteries of interest only, their expected Hounsfield Unit (HU) range was considered. In effect, we applied a windowing and converted all values of the volume of interest to values between 0 and 1. The windowing range is set between -50 HU and 450 HU, thus using a windowing with the following parameters 200 HU and 250 HU for the window center (WC) and the window length (WL), respectively.
\section{Methods}
The core part of the system is made of three consecutive steps. The localization of the slice containing the main PA bifurcation, the arteries segmentation and their respective measurement. The core system flow is shown in figure \ref{fig:system_design_pht}, and further described in details in this section, per component.

\subsection{Slice Selection}
\par In order to localize the slice containing the main PA bifurcation within the volume of interest, we trained a shallow CNN to recognize its properties. The CNN was given an equal number of slices suitable for the measurements and other randomly sampled slices from the volume of interest, in order to learn the differences between them.
Given a slice as input, the model returns a number between 0 and 1.
\par The CNN is composed of a few convolutions and max pooling layers, followed by a few fully connected layers. In effect, it consists of three 3x3 convolution-batch normalization-relu-max pooling blocks, followed by a fully connected-dropout-relu-fully connected block, with a sigmoid activation at the end. This model was trained using the binary cross entropy loss function. The CNN receives as input a resized slice of the volume of interest, as well as its two closest neighbors, namely the slices below and above. It then returns the score for the middle slice. The system chooses a unique slice by selecting the slice with the highest of such scores over the entire volume of interest.
\par This model was trained and validated on the 600 studies containing the exact measurements of the arteries. The ground truth was set to the slice measured by a single radiologist. Statistics on the test set showed that 78.8\% of the slices chosen by the network are within less than 5mm from the annotated scan and 94\% within 10mm. 

\subsection{Slice Segmentation}
\par Once a specific slice is selected, a fully convolutional CNN is used to segment both arteries. This network's architecture is a U-Net variant with skip connections, using VGG16-like \cite{Simonyan14} CNN as its encoder. We have limited it to three layers depth to avoid over-fitting of the model. In effect, this encoder is made of only two convolution-convolution-2x2 max pooling blocks followed by convolution-convolution-convolution-2x2 max pooling block, where all convolutions are 3x3, padded and contain rectified linear unit (relu) activations. Additionally, convolutions are added at the bottleneck of the encoder: a 7x7 convolution-relu-dropout and a 1x1 convolution-relu-dropout block. The up-sampling path consists of multiple transposed convolutions: a 1x1 convolution, followed by a 4x4 and a 16x16 transposed convolutions. A skip connection is used to make sure information from the down-sampling path is preserved in the up-sampling process. Indeed, the result of the first transposed convolution is combined with the result of the second convolutional block of the down-sampling path, by summing the resulting tensors. The up-sampling brings back the image to its original size. The model is trained with the sparse cross entropy loss.  
\par The segmentation data was built in an iterative process. During training, the model was fed with manually annotated masks and semi-automatic ones (only 15\% were manually tagged).
In effect, we gathered additional data for training by running the initial version of the model on unknown data set, and by visually reviewing the segmentation results. As input, the network takes three different windowed versions of the slice, in order to cope with various contrast phases of the scan. These slices are centered and zero-padded to fit a fixed 512x512 input. The middle slice uses the same windowing covering the wide range of common arteries HU intensities, so a WC of 200 HU, and WL of 250 HU. The second windowing is designed for delayed venous phase, and is defined with the following parameters: WC = 100 HU, WL = 150 HU. The third windowing is designed for an earlier phase of contrast, whereby both PA and Ao are enhanced and has the following parameters WC = 375 HU, WL = 175 HU.
The model returns a segmentation mask containing the three classes of interest Ao, PA and background.
\subsection{Arterial Diameter Measurement}
\par The last step in the system's flow is to measure the arteries diameter. Shape analysis and morphological operations are used in this component to measure them correctly. The segmentation from the previous step is checked, since it is expected to present a set of properties. In effect, the segmentation output is to contain exactly two distinguishable components, for each segmented artery. The Ao is to be circular, therefore its maximum eccentricity is set to $0.55$. The PA bounding box to filled area ratio should be limited, since a Y-shaped PA is expected. Also, the cross section size of both arteries is expected within reasonable boundaries of (256 to 500 $mm^2$), and the PA centroid is expected to the right of the Ao. These various checks caused approximately 21\% yield reduction, but led to higher performance and accuracy of the system.
\par Assuming a correct segmentation, the measurement process is started. Given that the Ao is assumed to be ellipsoidal, its measurement is relatively easy. Using the second order central moments to construct a covariance matrix of the image, one can fit an ellipse to its mask. We can then average the major and minor axis of the ellipse to provide an accurate estimate of its diameter. On the other hand, the PA can have very different shapes, and its measurement is therefore more complex. The main PA bifurcation is first localized on the skeleton of the PA binary mask using connected component analysis. Then, a stable interval of measurement above the bifurcation on the PA trunk is considered, at the same height as the Ao's center (following the measurement instructions given in section \ref{sec:intro}). By stable intervals here, we are looking into sections of the PA with minimal change in size, in order to measure properly measure the PA diameter. The interval values are finally averaged to compute the PA diameter.
\begin{figure}[t!]
% \centering
\hspace*{-1.7cm}
  \begin{subfigure}{.6\textwidth}
    \centering
    \includegraphics[width=.6\linewidth]{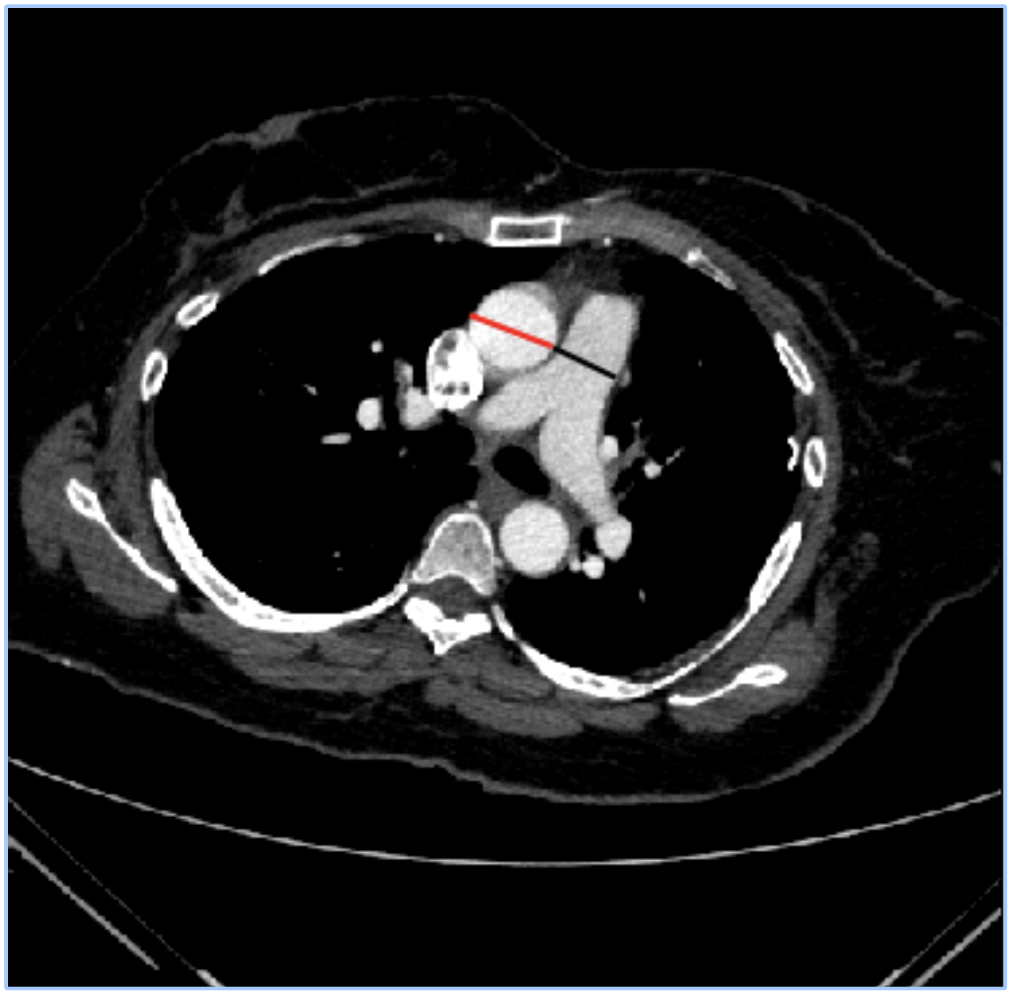}
    \captionsetup{justification=centering,margin=0.5cm}
    \caption{\label{fig:negative_pht}Case with a PA to Ao ratio $<$ 1.}
  \end{subfigure}%
  \begin{subfigure}{.6\textwidth}
    \centering
    \includegraphics[width=.6\linewidth]{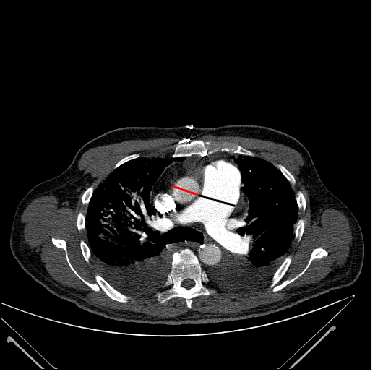}
    \captionsetup{justification=centering,margin=0.5cm}
    \caption{\label{fig:positive_pht}Case with a PA to Ao ratio $>$ 1.}
  \end{subfigure}%
\vspace*{0.4cm}
\captionsetup{margin=1cm}
\caption{\label{fig:sample_results}Sample of the system visual results. The measurements are shown in red and black for the Ao and PA, respectively.}
\end{figure}

\section{Experiments \& Results}
\begin{figure}
  \centering
  \includegraphics[width=0.7\linewidth]{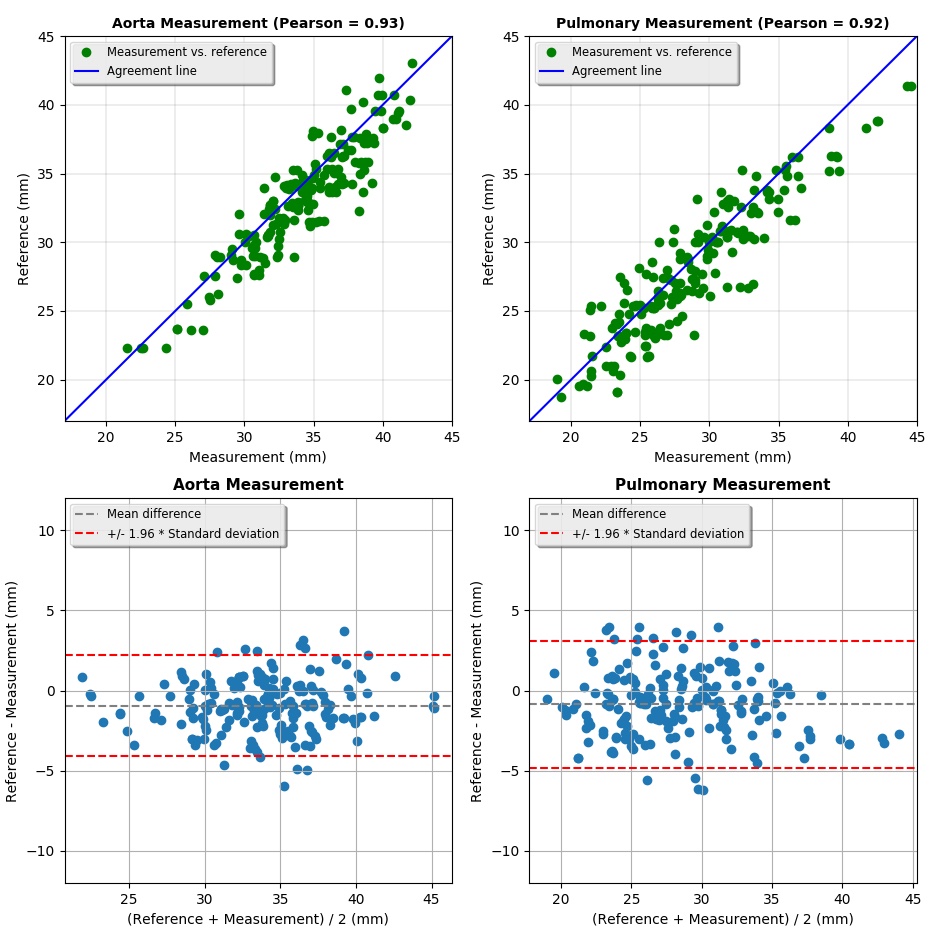}
  \captionsetup{margin=1cm}
  \vspace*{0.3cm}
  \caption{\label{fig:stats_pht} Our algorithm and the reference measurements have a high Pearson correlation score: 93\% for the Ao and 92\% for the PA. The top two plots show our measurements versus the reference per artery, knowing that perfect agreement is represented by the blue diagonal line. The bottom two plots display the Bland Altman plots per artery, with the mean difference plotted via the gray dotted line. The limits of agreement are also plotted in red (-4.10 mm, 2.22 mm) for the Ao and (-4.84 mm, 3.12 mm) for the PA.}
\end{figure}

Ground truth measurements were established by a radiologist on a test set of 288 CT studies, and compared to the algorithm outputs. Figure \ref{fig:stats_pht} presents multiple plots comparing the system's measurements to the ground truth reference measurements. Each compliant series within the study was automatically measured by the system. Cases with significant motion artifacts, or containing metallic artifacts were excluded from these experiments.
The results show Pearson correlation coefficient scores of 93\% for the Ao and 92\% for the PA. Bland-Altman plots showing the limits of agreement for the measurements of the two arteries are also presented. The Bland-Altman plots also show a mean difference and standard deviation of -0.94 mm and 1.6 mm for the Ao, and of -0.86 mm and 2.03 mm for the PA. Both negative biases are caused by the system tending to slightly overestimate the arteries diameter. This can be explained by the difficulty of the segmentation model to classify pixels at the arteries borders.
\par As demonstrated on the confusion matrix in table \ref{table:confusion_matrix_risk_stratification}, 91.9\% of the test cases were correctly risk stratified, based on the PA to Ao ratio $>$ 1 definition for PHT. Our solution's precision (positive predictive value) is 80.3\%, meaning that a patient with a positive algorithm answer has 80.3\% chance of having PHT. The system's sensitivity is 64.6\%, whereas its specificity reaches 97.0\%. The system's high specificity guarantees low false positive rates, thus making it ideal for screening tasks on large data sets.  
\par Over and above our results statistics, we asked radiologists to review the measurements of the algorithm as plotted on top of the measured axial slice on 410 additional test CT scans, and to label the results as acceptable or not. In addition, we sought to quantify how frequently the algorithm measured the arteries on the correct slice. The algorithm measurements were judged acceptable and in the correct location for 99.05\% and 99.76\% of the cases, respectively. The unacceptable cases included those with metallic artifacts or severe motion artifacts. These results demonstrate high agreement of the radiologists with the algorithm. A sample of the system visual results on cases with a physiologic and large PA to Ao ratio are shown in figure \ref{fig:sample_results}.
\par Concerning the performance, the system average run time per study was also evaluated. Using a Intel Xeon Processor E5-2695 v4 machine, which is a similar hardware to the one on radiologists workstation, the system run time is 70 seconds per study on average. Using a GTX 1081Ti Graphics Processing Unit, the system run time drops to 22 seconds per study on average.

\begin{table}
\setlength{\tabcolsep}{20pt}
\begin{center}
\begin{threeparttable}
\caption{Test set confusion matrix}\label{table:confusion_matrix_risk_stratification}
\begin{tabular}{lcccc}
\toprule
\toprule
& & \multicolumn{2}{c}{ Reference } \\
\rule{0pt}{3ex}  
& & no PHT & PHT\\
\midrule
\multirow{2}{*}{Algorithm}
& \quad no PHT & 81.7\% & 5.6\%\\
& \quad PHT & 2.5\% & 10.2\%\\
\bottomrule
\end{tabular}
\begin{tablenotes}[para,flushleft]
Confusion matrix of the algorithm versus the reference on the test set.
\end{tablenotes}
\end{threeparttable}
\end{center}
\end{table}

\section{Summary} The present solution provides automatic measurements of the Ao and PA from contrast-enhanced chest CT scans. It applies deep learning techniques as part of its flow in order to measure both arteries in the appropriate axial location. On top of the measurement values, the system also returns visual results, thus providing interpretable feedback to its operator. Evidence of PHT may be derived from the absolute PA diameter or the PA to Ao ratio, and detection of aortic aneurysms could be a secondary gain. Detecting PHT also allows more effective risk stratification of individuals with COPD, which could improve our ability to direct preventative and health surveillance resources. Thus preventing hospital admission and the clinical sequelae of COPD exacerbation. Knowing that radiologists usually do not measure both arteries while analyzing a CT scan, the described solution provides the operator with clinically valuable information with minimal additional time cost.
\par To the best of our knowledge, the system is the first such fully automated solution potentially allowing screening for COPD on large CT data sets. Future work will expand current techniques to apply to non-contrast CT cases.

\section{Acknowledgments} 
The authors would like to thank the Zebra Medical Vision team, especially Tomer Meir, Ayelet Akselrod-Ballin, Assaf Pinhasi, and Ronen Gordon for their helpful comments and discussions on this research, as well as Hadass Fishman for her help with the figures design.
% References
\bibliography{report} % bibliography data in report.bib
\bibliographystyle{spiebib} % makes bibtex use spiebib.bst

\end{document}